\documentclass[11pt]{article}
\usepackage{array}
\newcolumntype{L}[1]{>{\raggedright\arraybackslash}m{#1}}
\newcolumntype{C}[1]{>{\centering\arraybackslash}m{#1}}

\usepackage[final]{acl}

\usepackage{booktabs}
\usepackage{times}
\usepackage{latexsym}
\usepackage{graphicx}
\usepackage{subcaption}
\usepackage{float}
\usepackage[T1]{fontenc}

\usepackage[utf8]{inputenc}

\usepackage{microtype}
\usepackage{adjustbox}
\usepackage{inconsolata}

\usepackage{graphicx}

\usepackage{lipsum}
\usepackage{multirow}

\title{\textsc{SinFoS}: A Parallel Dataset for Translating Sinhala Figures of Speech}

\author{Johan Sofalas$^a$, Dilushri Pavithra$^a$, Nevidu Jayatilleke$^b$ \and Ruvan Weerasinghe$^a$ \\
  $^a$Research Department, 
  Informatics Institute of Technology, 
  Sri Lanka \\
  \texttt{\{johan.s, pavithra.r, ruvan.w\}@iit.ac.lk}, \\
  $^b$Department of Computer Science \& Engineering, University of Moratuwa, Sri Lanka \\
  \texttt{nevidu.25@cse.mrt.ac.lk}
  }

\begin{document}
\maketitle
\begin{abstract}
\textit{Figures of Speech} (FoS) consist of multi-word phrases that are deeply intertwined with culture. While \textit{Neural Machine Translation} (NMT) performs relatively well with the figurative expressions of high-resource languages, it often faces challenges when dealing with low-resource languages like Sinhala due to limited available data. To address this limitation, we introduce \textsc{SinFoS}, a dataset of 2,344 Sinhala figures of speech with cultural and cross-lingual annotations. We examine this dataset to classify the cultural origins of the FoS and to identify their cross-lingual equivalents. Additionally, we have developed a binary classifier to differentiate between two types of FoS in the dataset, achieving an accuracy rate of approximately 92\%. We also evaluate the performance of existing LLMs on this dataset. Our findings reveal significant shortcomings in the current capabilities of LLMs, as these models often struggle to accurately convey idiomatic meanings. 
By making this dataset publicly available, we offer a crucial benchmark for future research in low-resource NLP and culturally aware machine translation.
\end{abstract}

\section{Introduction}

Language and culture are deeply interrelated and significant mutual influence in multiple ways \cite{Hamidi}. FoS are the tools that make language expression more vivid, attractive, and effective \cite{lok}. They are built through a small set of meaning-construction mechanisms where speakers reuse familiar knowledge structures in new contexts \cite{DancygierSweetser2014Figurative}. Speakers utilise various figurative forms, such as exaggeration and idioms, as they often achieve discourse goals more effectively than literal words \cite{RobertsKreuz1994WhyFigurative}. While idioms are universal, each language features unique expressions with specific meanings, complicating the translation process and creating a sophisticated challenge \cite{medagama2021idiomatic}. 


The Sinhala language is part of the Indo-Aryan branch of the Indo-European language family with a rich and diverse literary heritage that has evolved over several millennia. It uses a unique script that is derived from the ancient Indian Brahmi script~\cite{jayatilleke-de-silva-2025-zero}. The origins of Sinhala can be traced back to between the 3rd and 2nd centuries BCE. Sinhala is the primary language of the Sinhalese people, who make up the largest ethnic group in Sri Lanka, and it is recognised as the first language (L1) for approximately 16 million individuals~\cite{de2025survey, jayatilleke2025sidiac}. According to the criteria established by ~\citet{ranathunga-de-silva-2022-languages}, Sinhala is classified as a lower-resourced language (Category 2).

Sinhala has a long and well-documented tradition of FoS (\raisebox{-0.5ex}{%
\includegraphics[height=1.4\fontcharht\font`\A,page=1]{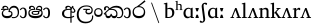}
}\hspace{-4.5pt}) that appears in both literary and everyday communication~\cite{senaveratne2005dictionary}. They emerged gradually as Sinhala speakers and writers needed brief ways to support religious, educational, and courtly objectives, communicate indirectly and memorably in everyday conversation, and enhance the aesthetic quality of their poetry \cite{Nawaz, Mieder}. Currently, Sinhala FoS are mainly preserved in collections such as books and dictionaries, with many manuscripts held by national institutions and temples \cite{Mieder}. In this study, we present \textsc{SinFoS}~\footnote{\scriptsize\url{https://huggingface.co/datasets/SloppyCalculator/SinFoS}}, the first Sinhala dataset of its kind with essential data to support the task of machine translation (target language: English).




\section{Related Works}
 A substantial body of research has examined FoS, including idioms \cite{sporlederidioms}, metaphors \cite{dodge-etal-2015-metanet}, proverbs \cite{bonin}, and other forms of figurative language \cite{kabra-etal-2023-multi}. 

\subsection{Existing FoS Corpora}

Resources are predominantly English-focused, whereas a smaller subset provides broader multilingual coverage, including European Portuguese, Danish, Chinese, and multi-language compilations such as \texttt{MABL} and \texttt{ID10M}~\cite{kabra-etal-2023-multi, tedeschi}. The datasets ranged in size from moderate idiom/proverb collections, small lexicons (hundreds to ~1,000 items) \cite{zhou-etal-2021-pie, LIDIOMS}, to (\~1,000–10,000) \cite{stowe-etal-2022-impli, Reddy}, with a few large-scale corpora (tens of thousands of instances/pairs or even larger textual corpora) \cite{ChID, Amsterdam}. Moreover, a limited number of datasets, such as \citet{adewumi-etal-2022-potential}, have a multi-phenomenon architecture that covers a greater variety of figurative categories, whereas many datasets are single-phenomenon resources that primarily target idioms or metaphors~\cite{sporlederidioms, dodge-etal-2015-metanet, IDEM, Konidioms}. 
 
\citet{Konidioms} introduce \texttt{KonIdioms}\footnote{\scriptsize\url{https://bit.ly/3Y4LGd3}}, an annotated Konkani idiom corpus (4,332 sentences and 817 potentially idiomatic expressions) designed to support automatic idiom identification and evaluation for this low-resource language. Furthermore, the \texttt{PARSEME} \footnote{\scriptsize\url{http://hdl.handle.net/11372/LRT-5124}} dataset release 1.3 provides multilingual annotations of \textit{Verbal Multiword Expressions} (VMWEs) across Arabic, Bulgarian, Chinese, Croatian, Greek, Hebrew, Hindi,
Irish, Latvian, Lithuanian, Maltese, Slovene, and Turkish languages, including a dedicated category for verbal idioms alongside other VMWEs types \cite{savary-etal-2023-parseme}. In contrast, the \texttt{SemEval-2022 Task 2 dataset}\footnote{\scriptsize\url{https://bit.ly/4s8k4Bt}} by \citet{madabushi2022semeval} focuses on idiomaticity-related modelling through sentence-level evaluation in English, Portuguese, and Galician, supporting tasks such as idiom detection and representation learning. Additionally, \texttt{IMIL}\footnote{\scriptsize\url{https://bit.ly/4p4SUsC}
}introduces an Idiom Mapping for Indian Languages resource that links idioms across Bengali, Gujarati, Hindi, Marathi, Punjabi, Tamil, and Urdu (with English mappings), enabling cross-lingual comparison and transfer for idiom processing \cite{agrawal-etal-2018-beating}.  

It is clear that datasets related to FoS are a significant area of focus for researchers in the field, including languages like Konkani~\cite{Konidioms}, which falls under the same language resource category (Category 2) as Sinhala~\cite{ranathunga-de-silva-2022-languages}. We have also discussed various existing FoS datasets for different languages in detail in Appendix~\ref{app: ExistingWork}.
 
\subsection{Classification of FoS}
Many studies have classified FoS into multiple categories, each supported by explicit definitions~\cite{BANOU}. \citet{Jang} categorised the \texttt{FLUTE} \footnote{\scriptsize\url{https://huggingface.co/datasets/ColumbiaNLP/FLUTE}} dataset into four categories, such as sarcasm, similes, idioms, and metaphors. Early work, such as the \textit{SemEval-2015 Task 11} by \citet{ghosh-etal-2015-semeval} and the discourse-oriented analysis by \citet{Musolff}, primarily focused on the interplay between sentiment and specific tropes, particularly irony, sarcasm, and metaphor, in social media and public discourse. Moreover, \citet{chakrabarty2021}  redefined figurative language data as instances of \textit{Recognising Textual Entailment} (\texttt{RTE}), structuring sentence pairs that comprise a premise and a hypothesis with an associated entailment label, by drawing on five pre-existing datasets (\texttt{Figurative-NLI}\footnote{\scriptsize\url{https://github.com/tuhinjubcse/Figurative-NLI}}~\cite{chakrabarty-etal-2020-generating}, datasets on \textit{irony} compiled by~\citet{van-hee-etal-2018-semeval}\footnote{\scriptsize\url{https://competitions.codalab.org/competitions/17468}} and \citet{ghosh-etal-2020-interpreting}\footnote{\scriptsize\url{https://bit.ly/44D3O1q}}, \texttt{Sarcasm SIGN}\footnote{\scriptsize\url{https://github.com/lotemp/SarcasmSIGN}}~\cite{peled-reichart-2017-sarcasm}, a metaphor dataset\footnote{\scriptsize\url{https://bit.ly/4rfWrGc}} by~\citet{chakrabartymermaid}) annotated for simile, metaphor, and irony, thereby constructing a corpus of more than \texttt{12,500 RTE} examples. \citet{Hayani} has classified the figurative texts into 12 categories, such as metaphor, personification, hyperbole, simile, metonymy, synecdoche, irony, antithesis, symbolism, and paradox.
 

\subsection{LLMs based Machine Translations}

As mentioned by \citet{Pramodya}, NMT systems for low-resource, morphologically rich languages such as Sinhala increasingly adopts transfer learning and fine-tuning of multilingual sequence-to-sequence LLMs rather than SMT. As mentioned by \citet{Thillainathan}, systematic pretraining on monolingual data followed by intermediate-task transfer provides better results than conventional single-stage fine-tuning of multilingual LLM-based MT systems in Sinhala-to-English translation. Despite these advancements, translating figurative language remains a challenging task. While retrieval-augmented prompting can improve the translation of idioms by offering helpful definitions or context~\cite{donthi2025improving}, comparative analyses show that, compared to human translations, outputs from LLMs often lack cultural nuance and tend to simplify creative metaphors~\cite{Sahari, Karakanta}.

Based on existing studies, it is evident that Sinhala figurative language is underexplored in the field of computational linguistics. Incorporating this resource by identifying the dominant semantic and cultural domains reflected in Sinhala figurative language, along with translating these data from Sinhala to English, will be significant for future research. Therefore, the purpose of this work is to present a dataset of Sinhala figurative language, capture its cultural nuances, and provide an essential resource for the task of machine translation from Sinhala to English.

\section{Data Collection and Annotation}

The \textsc{SinFoS} dataset consists of 2,344 unique FoS and was compiled from a carefully curated selection of authoritative resources, including various Sinhala literary works and selected Wikipedia entries. This section provides a detailed overview of the processes involved in assembling, annotating, and preprocessing the data.  An example of a record from the dataset that underwent these steps is illustrated in Figure~\ref{fig:record} in Appendix~\ref{app:dataset}. 

\subsection{Data Assembly}
\label{sec:data_assembly}

A significant portion of the data, approximately 65\%, was sourced from the prominent Sinhala books in this field. \raisebox{-0.5ex}{%
  \includegraphics[height=1.5\fontcharht\font`\A,page=1]{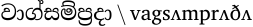}
} - Idioms~\cite{department_official_languages_idioms}, \raisebox{-0.5ex}{%
  \includegraphics[height=1.3\fontcharht\font`\A,page=1]{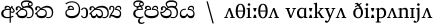}
} - Atheetha Wakya Deepanya \cite{senanayaka1880athetha}, and the Dictionary of Proverbs of the Sinhalese ~\cite{senaveratne2005dictionary}, while the remaining 35\% was extracted from Wikipedia~\footnote{\scriptsize\url{https://bit.ly/4qdZyO8}}.To ensure high fidelity to the source material, the core Sinhala expression was collected as the primary data entry. This is a foundational practice validated by benchmarks like the \texttt{IDIX}~\cite{sporlederidioms} and the \texttt{ChID}~\cite{ChID} corpora, which rely on the collection of specific linguistic expressions as the base unit for identification. 

\subsection{Annotation Process}

To ensure the accuracy of the sources, the annotation process closely followed the resources outlined in subsection~\ref{sec:data_assembly} and was carried out by native Sinhala speakers. Importantly, when primary sources lacked the expected information related to translations (although the attributes \textit{Literal / Visual Image} and \textit{Type of FoS} involved some human annotation as detailed in subsections~\ref{sec:t_fos} and~\ref{subsec:lit_vis}), the annotators refrained from using personal knowledge to avoid potential subjective interpretations. Instead, they strictly drew from previously verified resources. For example, \textit{What it really implies} was derived directly from the \textit{Corresponding FoS in English} found in the source books, utilising standard references such as Merriam-Webster~\cite{dictionary2002merriam} and the Cambridge Dictionary~\cite{brown2013cambridge} for validation. Similarly, missing \textit{Literal Image} entries were translated strictly from the FoS text, while \textit{Type of FoS} categories were assigned based solely on the logical frameworks outlined in subsection \ref{subsec:FoS_class} and Appendices \ref{app:ClassificationProverb}, \ref{appendix:strategies}. A final comprehensive review confirmed that all entries were grounded in these external standards, ensuring high data integrity. As a result of the procedures followed, certain records did not include some attributes, as shown in Table~\ref{tab:field_counts}.

\begin{table}[h!tb]
\centering
\renewcommand{\arraystretch}{0.9}
\resizebox{0.35\textwidth}{!}{
    \begin{tabular}{lr}
        \toprule
        \textbf{Attribute} & \textbf{Count} \\
        \midrule
        Sinhala (\raisebox{-0.4ex}{%
  \includegraphics[height=1.3\fontcharht\font`\A,page=1]{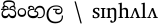}%
})      & 2344 \\
        Type of FoS & 2344 \\
        Literal / visual image & 2344 \\
        Corresponding FoS in English & 1571 \\
        What it really implies & 2059 \\
        Additional Context & 125 \\
        \bottomrule
    \end{tabular}
}
    \caption{Distribution of annotated fields in the dataset.}
    \label{tab:field_counts}
\end{table}

\subsubsection{\textit{Type of FoS}}
\label{sec:t_fos}
To clarify the figurative language associated with each record, the dataset includes a ``\textit{Type of FoS}'' attribute. This granularity was essential for determining the distinct processing strategies required for different figurative types, a necessity highlighted by the \texttt{PIE} corpus ~\cite{adewumi-etal-2022-potential}, which classifies data into specific types like metaphors and similes, and the \texttt{IMPLI} study \cite{stowe-etal-2022-impli}, which demonstrates that models process idioms and metaphors differently. 

\begin{table}[h!tb]
\centering
\resizebox{0.45\textwidth}{!}{
    \begin{tabular}{lr}
        \toprule
        \textbf{Type of FoS} & \textbf{Number of Entries} \\
        \midrule
        Proverbs (\raisebox{-0.5ex}{%
  \includegraphics[height=1.5\fontcharht\font`\A,page=1]{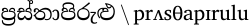}%
})      & 988 \\
        Idioms (\raisebox{-0.5ex}{%
  \includegraphics[height=1.5\fontcharht\font`\A,page=1]{Figures/wagwb.pdf}
}\hspace{-3pt})        & 1319 \\
        Adages (\raisebox{-0.5ex}{%
  \includegraphics[height=1.5\fontcharht\font`\A,page=1]{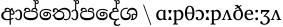}%
})       & 15 \\
        Idiosyncratic (\raisebox{-0.5ex}{%
  \includegraphics[height=1.5\fontcharht\font`\A,page=1]{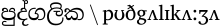}%
}) & 11 \\
        Sayings (\raisebox{-0.5ex}{%
  \includegraphics[height=1.5\fontcharht\font`\A,page=1]{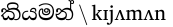}%
}\hspace{-3pt})       & 11 \\
        \bottomrule
    \end{tabular}
}
    \caption{Distribution of Entries by Figure of Speech Type.}
    \label{tab:fos_counts}
\end{table}

The entries are organised into five main categories, as detailed in Table~\ref{tab:fos_counts}. Most of the idioms were obtained from \cite{department_official_languages_idioms}, while the majority of the proverbs were gathered from \cite{senaveratne2005dictionary}. For certain FoS, specific types of FoS annotations were readily available, allowing us to directly categorise them within our classification strategy and document them accordingly. The remaining FoS were annotated based on the criteria outlined in subsection~\ref{subsec:FoS_class}. The guidelines provided in Appendix~\ref{appendix:strategies} were used to distinguish between proverbs and idioms. Additionally, proverbs were categorised into three subcategories based on their intent, origin, and conclusion. These annotations were performed according to the criteria in Appendix~\ref{app:ClassificationProverb}. Proverbs were assigned tags corresponding to the three categories mentioned earlier, while the other types of figurative speech were labelled directly, using their Sinhala names.

\subsubsection{\textit{Literal / Visual Image}}
\label{subsec:lit_vis}

\textsc{SinFoS} uses a ``\textit{Literal / Visual Image}'' annotation for each entry to provide a visual reference for non-native speakers by eliminating all abstract concepts, emotions, and symbolism. 
Documenting the literal imagery aligned with psycholinguistic research on imageability and methodologies for testing compositionality. Since the majority of these expressions are figurative, capturing the mental image was highly necessary. Furthermore, the inclusion of the implied meaning provided the ground truth required to test a model's ability to transcend surface definitions, mirroring the ``real vs. false definition'' methodology of the \textit{Danish Idiom Dataset} \cite{sorensen-nimb-2025-danish}.

Majority of the annotation was done using the above given sources as the relevant visual details were provided by them, whilst the others were annotated by translating the Sinhala FoS, word by word (e.g., \hspace{-3pt}\raisebox{-0.5ex}{%
  \includegraphics[height=1.5\fontcharht\font`\A,page=1]{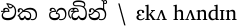}
}\hspace{-3pt} as ``With one voice''). The annotation process adhered to precise guidelines for aligning words, ensuring direct correspondence between the nouns and verbs in the original Sinhala text and their English descriptions. To maintain a ``Semantic Ground Truth'' and avoid introducing an outside context, only tangible objects and specific actions were documented. Furthermore, non-translatable ``cultural objects'' were preserved in their original form. For example \raisebox{-0.5ex}{%
  \includegraphics[height=1.5\fontcharht\font`\A,page=1]{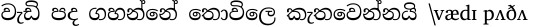}
} \raisebox{-0.5ex}{%
  \includegraphics[height=1.2\fontcharht\font`\A,page=1]{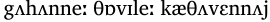}
} was annotated as ``Too much tom-toming means that the tovila is going to be spoilt'', retaining the word ``\raisebox{-0.5ex}{%
  \includegraphics[height=1.4\fontcharht\font`\A,page=1]{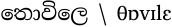}
} - Tovila (devil-ceremony, exorcism)''. This method helps prevent ``translation loss'' and ensures that the dataset’s literal accuracy is preserved, avoiding misleading interpretations that could arise from forced or inaccurate translations of culturally specific items.

\subsubsection{\textit{Corresponding FoS in English}}

The attribute ``\textit{Corresponding FoS in English}'' refers to the equivalent English figurative expression (FoS) for its Sinhala counterpart. One of the techniques explored by translators is direct substitution, which effectively facilitates the understanding of figurative language across different languages, even without explicit meanings~\cite{Adelnia2011TranslationOI}. This process further enabled the identification of cross-lingual equivalence and cultural parallels, a parallel alignment approach that was validated through the cross-linguistic mapping of proverbs in \textit{PROMETHEUS} \cite{ozbal-etal-2016-prometheus} and the alignment protocols of \textit{ParaDiom} \cite{ParaDiom}.

The FoS obtained from \citet{department_official_languages_idioms} included corresponding English FoS for all entries, whereas \citet{senaveratne2005dictionary} provided corresponding English FoS for only some entries, which were used for annotation. Additionally, the process of annotating this data also aided in determining the ``\textit{What it really implies}'' aspect for certain FoS.

\subsubsection{\textit{What it really implies}}

The ``\textit{What it really implies}'' column was established to clearly explain Sinhala figurative phrases in English, capturing their deeper meaning. It translates each Sinhala figurative expression into a shared human experience. Given that recognition of FoS is highly context-dependent, additional context is included to assist in disambiguation and cultural grounding. This field captures terms specific to Sinhalese culture, regional variations, and the folklore or stories behind specific figures of speech, ensuring the dataset serves as a comprehensive resource for understanding the ``naked truth'' behind the language. This is supported by the context-dependent annotation standards of \textit{EPIE} \cite{EPIE} and the cultural analysis frameworks of \textit{PROMETHEUS} \cite{ozbal-etal-2016-prometheus}.

To maintain clarity in the data and prevent lengthy explanations, the annotation process prioritised brevity over excessive detail. Only essential translations were included, omitting additional context or details that could complicate data analysis. Most implications in the expressions were derived directly from primary reference sources mentioned in the subsection~\ref{sec:data_assembly}. However, when a corresponding English equivalent was identified, the meaning was modified to align with the common interpretation of that English idiom. To guarantee reliable data, entries lacking a source-based explanation or an English equivalent were excluded. This mitigates the risk of inaccuracies or subjective misinterpretations. The annotations adhere to a specific format to aid in computational modelling. Behavioural advice and actions are expressed in the infinitive form. Character types or scenarios are described in formal terms. By eliminating secondary imagery and metaphorical elements, this approach clarifies the meaning for non-native speakers. It offers a clear ``ground truth '' for comparing the literal interpretation of a phrase with its actual significance.

\subsection{Data Pre-processing}
During the pre-processing stage, meticulous attention was devoted to punctuation, particularly in the context of FoS. The retention of punctuation marks in these instances is crucial, as they play a significant role in determining both prosody and syntactic structure, which are essential for achieving accurate processing. 
To ensure this dataset does not leak important information about figurative language, no further word-level or sentence-level filtration was conducted on any records, including those containing stereotypes, to facilitate authentic cultural analysis and the study of historical societal norms.

\section{Analysis of \textsc{SinFoS}}

The \textsc{SinFoS} dataset comprises 2,344 FoS, totalling 8,903 words. The literal image section includes 14,383 words, while the ``What it really implies'' section has 19,386 words. On average, each Sinhala FoS consists of 3.798 words. A brief overview of the dataset statistics is shown in Table \ref{tab:summary_stats}.

\begin{table}[h]
\centering
\resizebox{0.95\columnwidth}{!}{%
\begin{tabular}{lccccc}
\toprule
\textbf{Category} & $\mathbf{N}$ & \textbf{Mean} &  \textbf{Median} & \textbf{Max} & \textbf{Total} \\
\midrule
\textbf{Sinhala FoS} & \textbf{2344} & \textbf{3.80} & \textbf{3} & \textbf{24} & \textbf{8903} \\
Literal / visual image & 2344 & 6.14 & 5 & 38 & 14383 \\
What it really implies & 2059 & 9.41 & 8 & 56 & 19366 \\
Corresponding FoS in English & 1571 & 3.44 & 3 & 21 & 5401 \\
\bottomrule
\end{tabular}%
}
\caption{Summary statistics of word counts across different categories.}
\label{tab:summary_stats}
\end{table}

\subsection{Classification of FoS}
\label{subsec:FoS_class}

The classification of Sinhala FoS (\raisebox{-0.5ex}{%
  \includegraphics[height=1.5\fontcharht\font`\A,page=1]{alankara.pdf}
}\hspace{-4.5pt}) is complex due to the fluidity of the language and its deep rooting in oral tradition. As mentioned in the subsection~\ref{sec:t_fos}, this study classified Sinhala FoS into five main categories. The etymological roots of these terms provide a necessary framework for understanding their usage.

\begin{figure}
    \centering
    \setlength{\fboxsep}{1pt}
    \setlength{\fboxrule}{0.4pt} 
    \fbox{
        \includegraphics[width=0.85\linewidth]{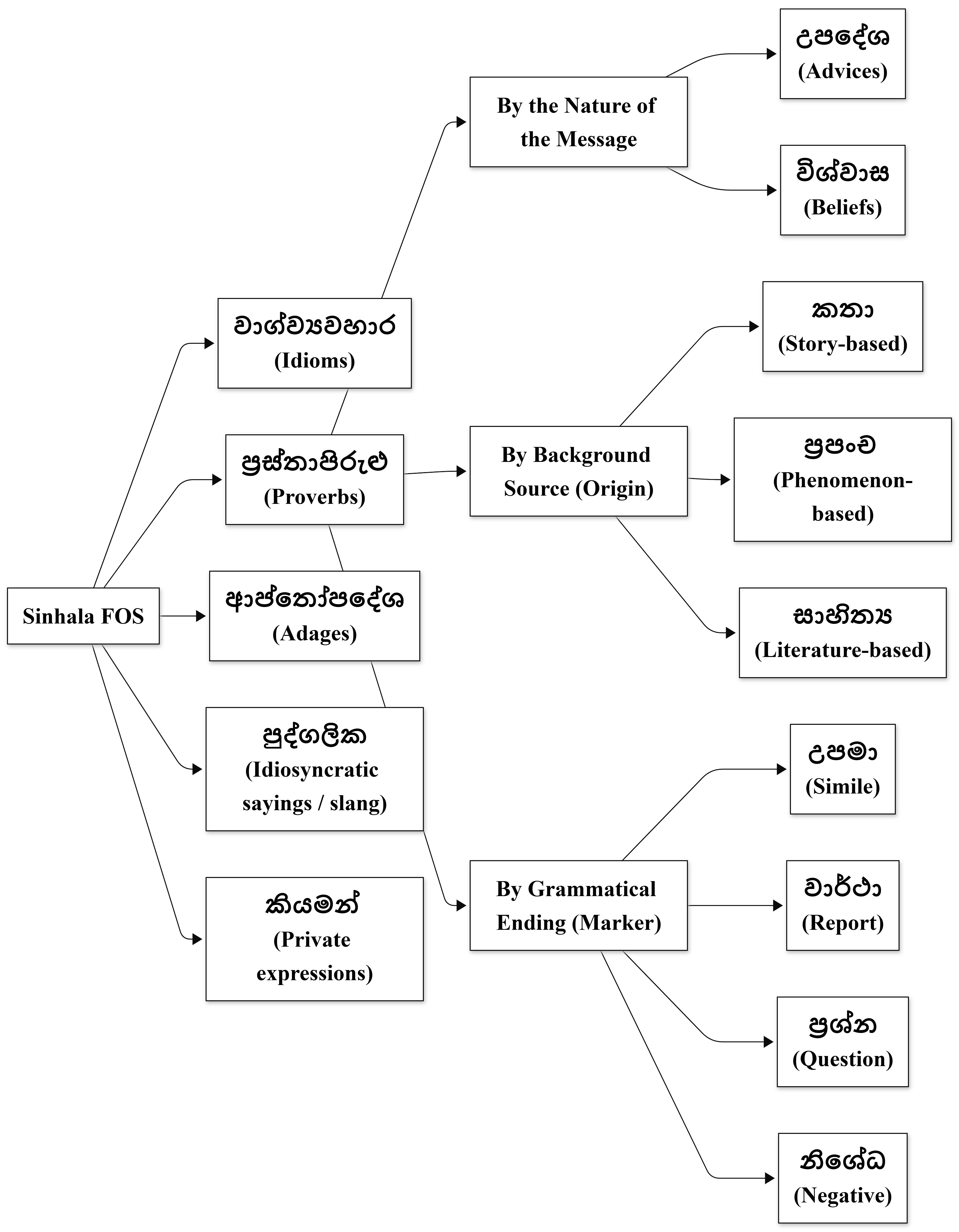}
    }
    \caption{Summary of Sinhala FoS Dataset Classification}
    \label{fig:classification}
\end{figure}

\vspace{0.5em}
\noindent
  \hspace*{-3pt}%
   \raisebox{-0.5ex}{ 
\includegraphics[height=1.5\fontcharht\font`\A]{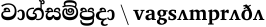}}
  \ \textbf{(Sinhala idioms):} Derived from the Sanskrit roots
``\raisebox{-0.5ex}{ 
\includegraphics[height=1.5\fontcharht\font`\A]{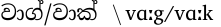}}'' (speech/word) and ``\raisebox{-0.5ex}{ 
\includegraphics[height=1.5\fontcharht\font`\A]{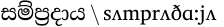}}'' (tradition/heritage), this term refers to speech patterns established by long-standing usage. Unlike proverbs, which are often wisdom-based, these are usage-based constructs where the meaning transcends the literal definitions of the individual words.
These are typically incomplete phrases or fragments, often ending in a verb. For example\raisebox{-0.5ex}{ 
\includegraphics[height=1.5\fontcharht\font`\A]{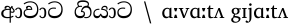}} literally translates to ``For coming and going'' while it actually means ``not friendly, and showing little interest in other people in a way that seems slightly rude''.

\vspace{0.5em}
\noindent
  \hspace*{-3pt}%
  \raisebox{-0.5ex}{ 
\includegraphics[height=1.5\fontcharht\font`\A]{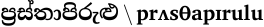}}
  \ \textbf{(Sinhala proverbs):} This is a compound of
``\raisebox{-0.5ex}{ 
\includegraphics[height=1.5\fontcharht\font`\A]{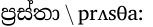}}''
denoting a specific occasion, moment, or opportunity, and
``\raisebox{-0.5ex}{ 
\includegraphics[height=1.5\fontcharht\font`\A]{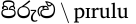}}''
referring to a simile, reply, or adage. Consequently, this functions as a
``situational simile'', a pre-packaged linguistic unit invoked to address
a specific incident by comparing it to a known truth. In contrast to Sinhala idioms, Sinhala proverbs are syntactically complete
sentences or clauses that can stand alone. For example
\hspace*{-3pt}\raisebox{-0.5ex}{ 
\includegraphics[height=1.6\fontcharht\font`\A]{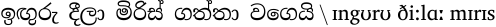}}
\hspace*{-3pt}\raisebox{-0.5ex}{ 
\includegraphics[height=1.3\fontcharht\font`\A]{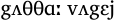}}
(Like exchanging ginger for chili). To provide a granular analysis, Sinhala proverbs were further classified based on the nature of the message, the source of the background, and the grammatical ending as mentioned in Appendix~\ref{app:ClassificationProverb}.

\vspace{0.5em}
\noindent
  \hspace*{-3pt}%
   \raisebox{-0.5ex}{ 
\includegraphics[height=1.5\fontcharht\font`\A]{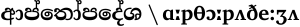}}
  \ \textbf{(Sinhala adages):} Unlike figurative proverbs, these are literal directives. They represent
the prescriptive aspect of the language (what one should do), distinct
from the descriptive nature of idioms. An example of adages in \textsc{SinFoS} is \raisebox{-0.5ex}{\includegraphics[height=1.45\fontcharht\font`\A]{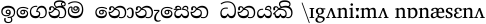}}  \hspace*{-4pt}\raisebox{-0.5ex}{ 
\includegraphics[height=1.45\fontcharht\font`\A]{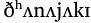}} (Education is an indestructible form of wealth).

\vspace{0.5em}
\noindent
  \hspace*{-3pt}%
  \raisebox{-0.5ex}{ 
\includegraphics[height=1.5\fontcharht\font`\A]{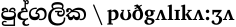}}
  \ \textbf{(Idiosyncratic):} These are hyper-local sayings used by individuals or small groups. While
not yet FoS in the public domain~\cite{crocker1977social}, they represent the genesis point of language evolution, where personal metaphors potentially graduate into
public idioms over time. Slang also falls under this category. For example the phrase \hspace*{-3pt}\raisebox{-0.5ex}{ 
\includegraphics[height=1.5\fontcharht\font`\A]{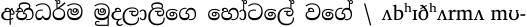}} \raisebox{-0.5ex}{ 
\hspace*{-3pt}\includegraphics[height=1.2\fontcharht\font`\A]{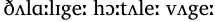}} (Like Abhidharma mudalali's hotel) would be well understood by the people living in the surroundings but not by everyone.

\vspace{0.5em}
\noindent
  \hspace*{-3pt}%
  \raisebox{-0.5ex}{ 
\includegraphics[height=1.5\fontcharht\font`\A]{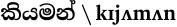}}%
  \ \textbf{(Sinhala sayings):} Concise verbal phrases are commonly used in daily conversation to express a thought, comment, or observation. In contrast to proverbs or idioms, these do not inherently possess a moral lesson, universal truth, or established figurative interpretation recognised by a large group. As an example~\raisebox{-0.5ex}{ 
\includegraphics[height=1.4\fontcharht\font`\A]{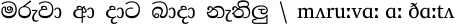}}
\hspace*{-4pt}\raisebox{0.1ex}{ 
\includegraphics[height=1\fontcharht\font`\A]{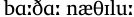}} (When death comes, there is no let or hindrance).

\begin{table}[htbp]
\begin{center}
\resizebox{0.95\columnwidth}{!}{%
\begin{tabular}{l l c c c}
\hline
\textbf{Model} & \textbf{Vectorization} & \textbf{Accuracy} & \textbf{P-Rec.} & \textbf{I-Rec.} \\
\hline
Gaussian Naive-Bayes & Word2Vec & 90.34\% & 92\% & 89\% \\
LinearSVC & Word2Vec & 90.34\% & 90\% & 91\% \\
Random Forets & TF-IDF (Char 3) & 89.27\% & 83\% & 94\% \\
XG-Boost & TF-IDF (Char 3) & 90.13\% & 86\% & 93\% \\
Ensemble (SVC+RFM+XGB) & TF-IDF (Char 3) & 90.56\% & 85\% & 95\% \\
Bi-LSTM & - & 91.63\% & 86\% & 96\% \\
Deep NN & - & 92.7\% & 94\% & 92\% \\
\hline
\end{tabular}%
}
\caption{Model Performance Comparison. Further details in Appendix~\ref{class_performance}. *Note that P-Rec = Recall for Proverbs and I-Rec = Recall for Idioms.} 
\label{tab:model_performance}
\end{center}
\end{table}

The dataset primarily consists of Sinhalese proverbs and idioms, leading to the creation of a binary classification model aimed at distinguishing between proverbs and idioms. A Voting Ensemble model, incorporating Support Vector Machines (SVM), Random Forest, and XGBoost with TF-IDF Character 3-Gram vectorisation, achieved an impressive accuracy of 90.56\%. This approach, based on character-level processing, effectively tackled the intricacies of Sinhala morphology~\cite{priyanga2017sinhala} by detecting sub-word elements rather than relying solely on exact phrases. The implementation of Word2Vec embeddings significantly improved performance compared to experiments based on TF-IDF (sparse vector representation). This includes the accuracy of the TF-IDF Character 3-Gram in both the Gaussian Naive Bayes and Linear SVC models, achieving an accuracy of 90.34\% in each case. The analysis indicated that specific verb endings served as strong indicators of idiomatic expressions, while comparative particles and rhythmic consonant clusters were associated with proverbs. Incorporating 3-gram TF-IDF was used to leverage the identified patterns, resulting in models with these embeddings performing better than their word-level counterparts. The semantic understanding provided by dense embeddings, such as Word2Vec, also proved effective in recognising these patterns. Ultimately, utilising a Deep Feed Forward Neural Network (Deep NN), which offers superior semantic understanding, achieved the highest overall accuracy of 92.7\% and the best recall for proverbs at 94\%. The embeddings for the LSTM and Deep NN models are not specified in Table~\ref{tab:model_performance}, as they relied on the standard TensorFlow Keras embeddings that learned directly from the training data.

\subsection{Cultural Analysis}

This research employed a hybrid methodological approach that combined both inductive and deductive thematic analysis to explore the relationship between physical imagery and cultural significance in Sinhala FoS. This computational analysis was conducted on English translations of the dataset. The analysis identified two main aspects of the FoS: ``\textit{Literal / Visual Image}'' (Source Domain), which consists of the tangible visual components that make up the figure of speech, and ``\textit{What it Really Implies}'' (Target Theme), which signifies the deeper abstract or cultural meanings conveyed by the text.
To minimise researcher bias and ensure that the coding frameworks were derived from raw data rather than from preconceived notions, we emphasised a bottom-up discovery phase. This inductive stage employed unsupervised machine learning methods to uncover naturally occurring patterns. Specifically, we applied TF-IDF vectorisation (using unigrams and a maximum of 2,000 features) along with K-Means clustering (k=5) to analyse the ``What it Really Implies'' dimension and uncover hidden linguistic clusters.

Additionally, we conducted a frequency analysis using a \textit{Bag-of-Words} (BoW) model for both the ``\textit{Literal / Visual Image}'' and ``\textit{What it Really Implies}'' dimensions. This analysis allowed us to identify the most frequent and significant terms in each cluster, categorising specific words under different themes and establishing a data-driven basis for the theoretical coding frameworks.
After completion of the exploratory phase, the recognised patterns were compiled into an organised dictionary for the deductive phase. We employed a rule-based classification system, using the specific keywords identified in the earlier phase as indicators of broader cultural categories. The algorithm compared the text against this predefined dictionary; if a keyword associated with a certain category was found, that category was assigned to the entry. This approach enabled multi-label classification, assuming that the subject matter remained consistent across the figurative language, thereby confirming that the detected keywords were suitable representations of the main concepts. 

Lastly, a bivariate cross-tabulation was performed to quantitatively evaluate the connections and dependencies between the identified Source Domains and Target Themes.
The findings reveal that Somatic (Body) and Agrarian (Nature) imagery are the most prevalent source domains, with notable mentions of the hand (n=56), water (n=46), and trees (n=43). The most frequently encountered themes are Ethics \& Moral Character (n=162) and Karma \& Consequence (n=127). This suggests a distinct metaphorical framework in which nature-related metaphors primarily promote moral conduct (n=20), while physical imagery specifically illustrates the tangible repercussions of karmic consequences (n=14). The distribution of literal source domains and abstract cultural themes observed in \textsc{SinFoS} is summarised in Table~\ref{tab:domain_occurrences} in Appendix~\ref{app:cultural_analysis}. This implies that these FOS primarily serve as mechanisms for reinforcing social norms rather than simply providing descriptive observations.

\subsection{Cross-Lingual Equivalence Analysis}

This study investigates a collection of 1,571 Sinhala phrases that have English ``Literal/ Visual Image'' translations. This sample is derived from the initial dataset of 2,344 phrases, as the remaining 773 lack direct English equivalents. The findings indicate a notable cultural divergence, demonstrated by a symbolic overlap score of merely 0.05 using the Jaccard Index and a lexical similarity score of 0.32. The lexical similarity was calculated using the sequence matcher in the \texttt{difflib}~\footnote{\scriptsize\url{https://bit.ly/4p48y7o}} library, which employs the \textit{Ratcliff/Obershelp Algorithm}~\cite{ratcliff1988pattern}. This implies that although the functional meanings align, the underlying metaphors originate from distinct contexts. 

For example, Sinhala employs the expression ``exchanging ginger for chilli,'' while English phrases refer to ``jumping out of the frying pan into the fire.'' In terms of structure, 93.3\% of the phrases retain their original form, while 4.9\% transition from Sinhala similes into English metaphors. An illustration is ``Like the eye,'' which transforms into ``Apple of one’s eye.'' 

Furthermore, expressions in Sinhala are, on average, 32\% longer than their English counterparts, yielding a ratio of 1.32. This distinction is effectively showcased by the English phrase ``red herring,'' which in Sinhala translates to an elaborate depiction where "the fox conceals the fowl in the forest and scurries about, swinging a coconut husk from its mouth."

\section{Benchmarking on LLMs}

In this section, we use \textsc{SinFoS} as a benchmark to evaluate the performance of selected LLMs and \textit{Small Language Models} (SLMs) in translating these complex expressions. A subset of 499 FoS was curated based on specific criteria: they represent diverse categories and possess intricate meanings that are particularly challenging for models to interpret \cite{madabushi2022semeval}. To ensure a comprehensive evaluation, we employed stratified sampling, purposefully oversampling rare categories, such as adages (11), ``private'' expressions (10), and sayings (3), which are often overshadowed by dominant idioms (190) and proverbs (285). This approach allows for a robust assessment of model capabilities across the full spectrum of figurative language, prioritising interpretative difficulty to test the distinction between literal cues and cultural nuances \cite{madabushi2022semeval}. Furthermore, proverbs were broken down into their core elements (story, nature, and literature) to better analyse the depth of cultural understanding.

\begin{figure}
    \centering
    \setlength{\fboxsep}{1pt} 
    \setlength{\fboxrule}{0.4pt} 
    \fbox{%
        \includegraphics[width=0.9\columnwidth]{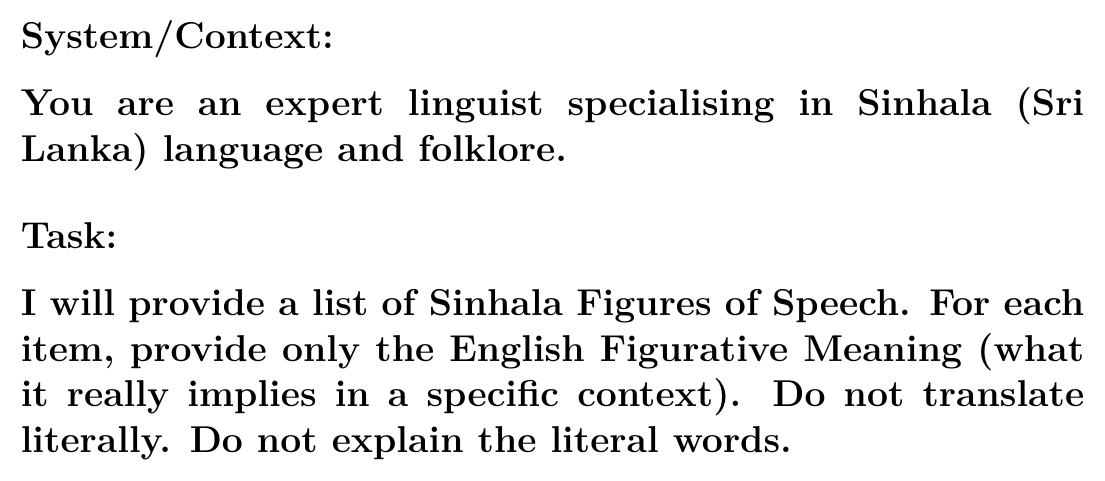}%
    }%
    \caption{Prompt used to generate responses from LMs.}
    \label{fig:prompt}
\end{figure}

We used the same prompt for all models to establish a consistent evaluation baseline. Figure~\ref{fig:prompt} shows the prompt provided to the \textit{Language Models} (LMs) to elicit the meanings of the FoS. This method helps avoid prompt-induced bias, as small variations in wording could unintentionally favour one LM over another, ensuring that the responses are directly comparable.

\begin{table}[htbp]
\centering
\resizebox{0.7\columnwidth}{!}{%
    
    \begin{tabular}{l|c|c}
        \hline
        \textbf{Model} & \textbf{Cosine} & \textbf{Fidelity} \\
         & \textbf{Similarity} & \textbf{Scores} \\
        \hline
        \texttt{Gemini 3 Pro}     & 0.6678 & 0.3117 \\
        \texttt{Llama 4 Maverick} & 0.6400 & 0.2351 \\
        \texttt{Grok 4.1}         & 0.6354 & 0.2361 \\
        \texttt{GPT 5.2}          & 0.6221 & 0.2090 \\
        \texttt{DeepSeek-V3.2}    & 0.6126 & 0.2052 \\
        \texttt{Claude Sonnet 4}  & 0.5972 & 0.1628 \\
        \texttt{Gemma}            & 0.6024 & 0.1914 \\
        \texttt{GPT 4.1 mini}     & 0.5816 & 0.1300 \\
        \texttt{Qwen 3}           & 0.5596 & 0.1247 \\
        \hline
    \end{tabular}
    }
    \caption{Performance of language models on Sinhala FoS.}
    \label{tab:model_accuracy}
\end{table}


\begin{figure*}[ht]
    \centering
    \begin{subfigure}{0.49\textwidth}
        \centering
        \includegraphics[width=0.8\linewidth]{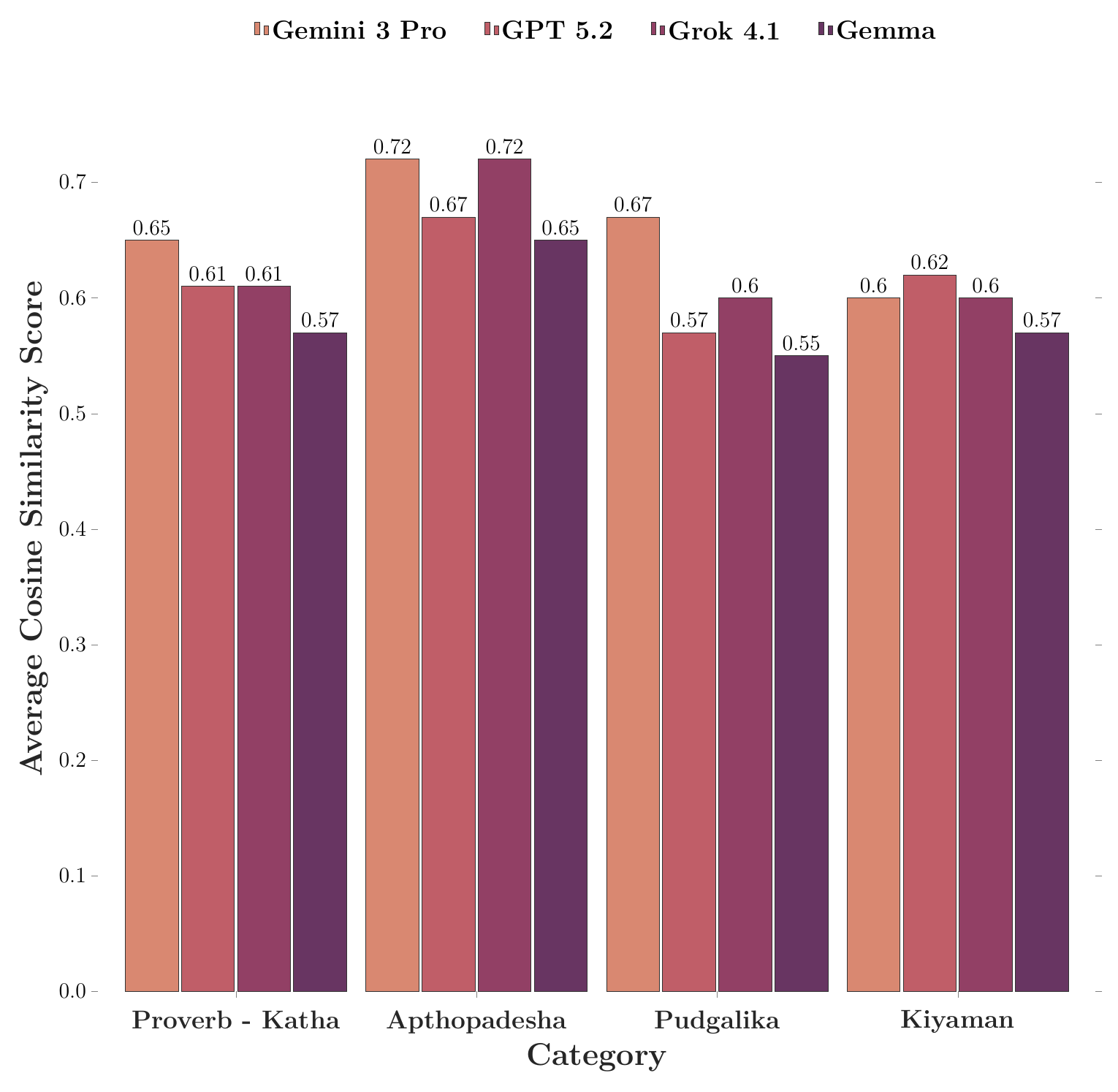}
        \caption{Cosine Similarity Score Comparison for Selected Categories}
        \label{fig:similarity}
    \end{subfigure}
    \hfill 
    \begin{subfigure}{0.49\textwidth}
        \centering
        \includegraphics[width=0.8\linewidth]{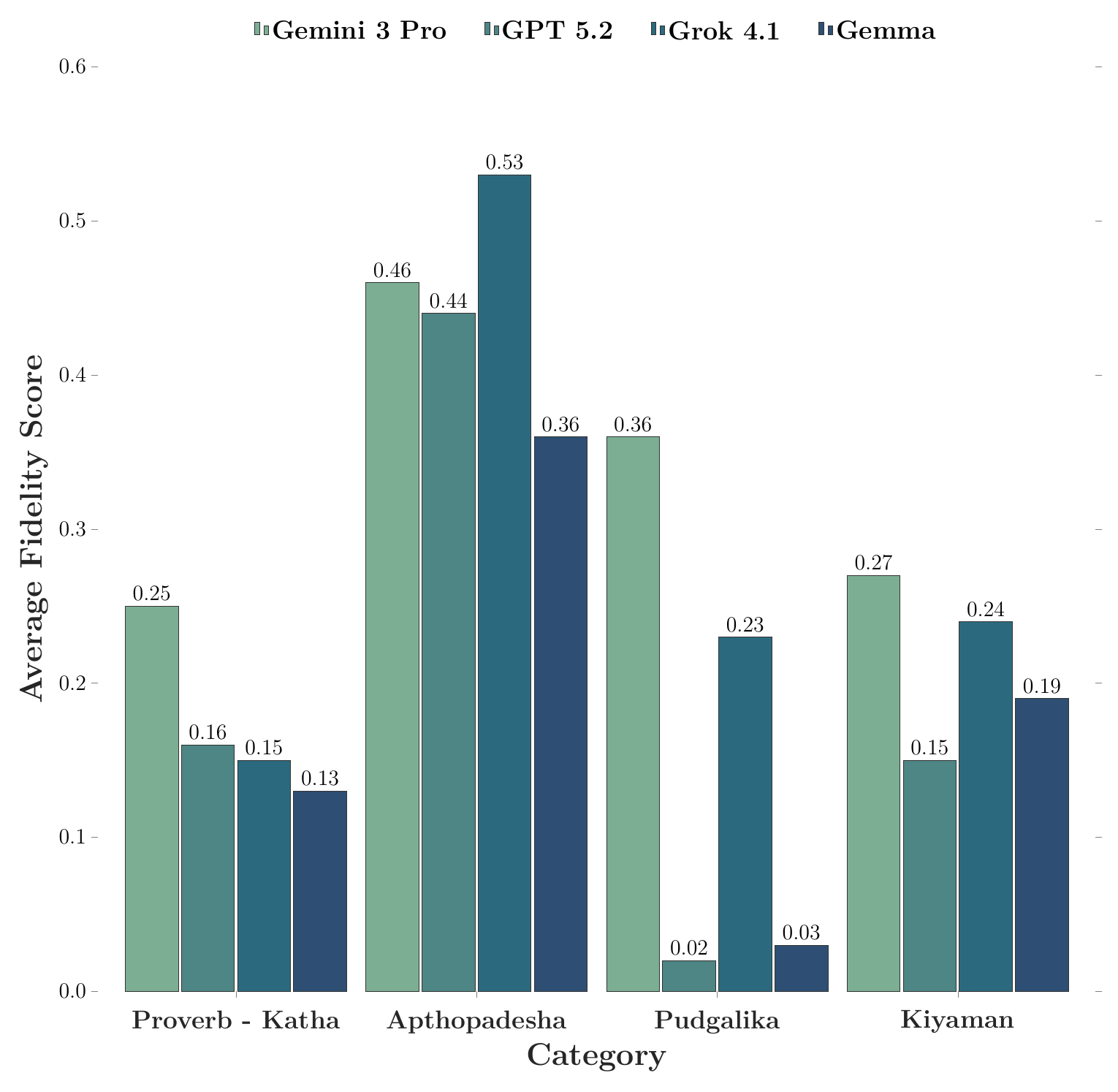}
        \caption{Fidelity Score Comparison for Selected Categories}
        \label{fig:fidelity}
    \end{subfigure}
    \caption{Benchmarking LLM Performance: (a) Cosine Similarity and (b) Fidelity Score. Information on all LLM performances could be found in Appendix~\ref{llm_performance}.}
    \label{fig:benchmarks}
\end{figure*}

To evaluate how effectively LMs grasp FoS in Sinhala, this research employs a dual framework that examines both context retrieval and logical comprehension. This method reflects the two-step process of theme identification and truth condition mapping by~\citet{reimers-gurevych-2019-sentence}. The initial phase utilises a bi-encoder architecture with \texttt{FlagEmbedding} (specifically the \textit{BAAI/bge-large-en-v1.5}~\footnote{\scriptsize\url{https://huggingface.co/BAAI/bge-large-en-v1.5}} model) to calculate Cosine Similarity between the outputs of the model and the meanings annotated in the dataset. This model was selected for its state-of-the-art performance on the \textit{Massive Text Embedding Benchmark} (MTEB), ensuring precise high-dimensional mapping that outperforms standard baselines in capturing ``Semantic Relatedness'' \cite{chen-etal-2024-m3, madabushi2022semeval}.

Although this segment efficiently penalises thematic discrepancies, such as mixing ``betrayal'' with ``love,'' it may be influenced by the ``Keyword Bag'' problem, in which comparable terms obscure gaps in logical coherence. For example, the idiom \raisebox{-0.5ex}{%
  \includegraphics[height=1.5\fontcharht\font`\A,page=1]{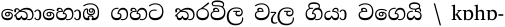}
} \hspace*{-1.5pt}\raisebox{-0.5ex}{%
  \includegraphics[height=1.2\fontcharht\font`\A,page=1]{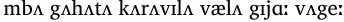}
} which implies the compatible union of two negative forces (literal image: `like the karawila creeper twining round the kohomba tree') received a high similarity score of 0.805 for the \texttt{DeepSeek V3} translation, 'a mismatched or absurd pairing', despite the model's output conveying the exact opposite meaning. 

To tackle this issue, the second segment measures the Fidelity Score, which implements a Cross-encoder (\textit{stsb-roberta-large}) to evaluate intricate dependencies by analysing sentences concurrently \cite{reimers-gurevych-2019-sentence}. In this context, Fidelity represents the semantic faithfulness of the model's output to the ground truth. This functions as a replacement for ``Semantic Entailment,'' aiding in the differentiation between sentences that share similar phrasing but convey distinct meanings, such as ``the dog bit the man'' versus ``the man bit the dog'' \cite{li2024idiomkb}. By utilising the full self-attention mechanism of the Cross-encoder, the framework captures the syntactic nuances often missed by Bi-encoder models. Integrating this Fidelity Score with the first segment provides robust safeguards against ``Low-Resource Hallucination,'' enabling a comprehensive assessment of Language Models in the Sinhala language \cite{benkirane-etal-2024-machine}. 

At the same time, the Fidelity scores struggle with something known as the ``Hyper-Literal'' problem, where creative paraphrasing could be penalised. For example, the phrase \raisebox{-0.5ex}{%
  \includegraphics[height=1.5\fontcharht\font`\A,page=1]{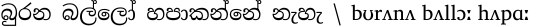}
} \raisebox{-0.01ex}{%
  \includegraphics[height=1\fontcharht\font`\A,page=1]{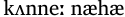}
} is directly translated as ``Barking dogs don't bite'' by DeepSeek V3. In the case of translating FoS, substitution with a valid FoS is considered to be a valid form of translation \cite{Adelnia2011TranslationOI}, but Fidelity gives it a modest score of 0.0089, as both phrases do not have lexical overlap. Relying only on one of these metrics can cause blind spots and skew evaluation results.


Therefore, by including both metrics, we can better assess the model's performance. This method identifies ``hallucinated relevance,'' where high Cosine scores suggest understanding, but low Fidelity scores indicate a lack of grasp on underlying intent. This helps benchmark true understanding over mere statistical matching. Table \ref{tab:model_accuracy} displays the average Cosine Similarity Scores and average Fidelity Scores obtained by each of these models across all the FoS available on the stratified sample obtained on the dataset based on the types of FoS, difficulty and figurativeness. 

The assessment of nine advanced models reveals that \texttt{Gemini} stands out in its ability to analyse Sinhala FoS, achieving the top scores in Cosine Similarity and Fidelity. The success of the smaller \texttt{Gemma} model indicates that cultural relevance takes precedence over the model size. Nonetheless, there is an issue known as the ``illusion of competence.'' Some models can effectively retrieve context but falter in logical comprehension. As a result, they may identify the correct domain but often misinterpret the meanings. Conventional metrics, such as BLEU, do not adequately address this challenge. Furthermore, models such as \texttt{GPT-4o mini} and \texttt{Qwen} exhibit ``broken figurative triggers,'' offering literal interpretations instead of figurative ones for specific expressions. While most models perform well with sayings that align with Western proverbs, they tend to struggle with distinct and folklore-inspired proverbs. This stems from their literal approach to translation, which neglects the cultural context needed to understand nuances.

\section{Conclusion}

This study introduces \textsc{SinFoS}, a dataset containing 2,344 Sinhala FoS accompanied by expert-verified explanations. The annotation process is comprehensively explained in the paper. The available details were entered into the dataset, and the missing details were handled in a manner consistent with the structure of the entered details to ensure the dataset's accuracy and validity.

The analysis of the dataset emphasises a significant disparity in meaning between Sinhala idioms and their English equivalents. The cross-linguistic examination revealed the disparities among the languages, while the cultural analysis showcased the distinct culture reflected in the FoS, emphasising the challenges of translation. While LLMs can effectively handle FoS with direct English translations, they often struggle with culturally specific terminology. This can result in inaccuracies or literal conversions. Future research should focus on improving the verification of these results by implementing ablation studies and presenting statistical significance. Consequently, \textsc{SinFoS} serves as a vital resource for developing novel approaches in Machine Translation and modelling frameworks that seek to integrate cultural insights into languages with fewer resources. 

\section*{Limitations}

\paragraph{Sinhala meaning unavailability:}
A key limitation of this study is the incomplete availability of English meanings for some Sinhala FoS. In several cases, authoritative definitions or consensus interpretations were not available in accessible reference sources, which constrained some of the analysis, such as where cross-lingual analysis could not be performed across all the FoS, and the domains spoken by these FoS could not be analysed in the cultural analysis.

\paragraph{Meaning loss in English rendering:}
Some Sinhala FoS are highly culture-bound, context-dependent, or rely on implicit background knowledge, making direct English rendering difficult and increasing the risk of ambiguity or meaning loss. As a result, a portion of the dataset may contain paraphrased or approximate meanings rather than fully equivalent English interpretations, which can affect translation quality and downstream classification performance.

\paragraph{Class imbalance in 
    \raisebox{-0.5ex}{
  \includegraphics[height=1.1 em,page=1]{Figures/pudgalika.pdf}%
} and
    \raisebox{-0.5ex}{%
  \includegraphics[height=1.1 em,page=1]{Figures/kiyaman.pdf}%
} categories:}
The dataset exhibits class imbalance, particularly within the \hspace*{-3pt}\raisebox{-0.5ex}{
  \includegraphics[height=1.1 em,page=1]{Figures/pudgalikawb.pdf}
} and \raisebox{-0.5ex}{
  \includegraphics[height=1.1 em,page=1]{Figures/kiyamanwb.pdf}
} categories, where only 11  instances were available for both categories. Therefore, the analysis done was heavily influenced by the dominant idioms and proverbs. A classification model could not be trained to classify all FoS due to the class imbalance.

\bibliography{references, anthology-1, anthology-2}

\appendix

\section{Existing Datasets Utilised} \label{app: ExistingWork}

\subsection{Germanic-Language Corpora}
\citet{sporlederidioms} have introduced \texttt{IDIX} dataset which contains English idioms. In there, they have mentioned idioms as a contextual disambiguation problem. Rather than focusing on token-level labels, \citet{haagsma-etal-2020-magpie} emphasises an inventory of potentially idiomatic expression types in English, that may be idiomatic depending on usage. The \texttt{PIE} dataset presented by \citet{zhou-etal-2021-pie} has been constructed to aid in the analysis of idiom paraphrasing by connecting idiomatic statements to alternatives that preserve meaning. \texttt{PIE} dataset by \citet{adewumi-etal-2022-potential}, constructed from \texttt{BNC} and \texttt{UKWaC}, provides an additional comprehensive English-only structure where instances are labelled across different FoS, such as metaphor, simile, euphemism, and irony, alongside literal examples. This extends beyond a binary idiom/literal structure to facilitate fine-grained multi-class categorisation of figurative language. 

The \texttt{VU Amsterdam Metaphor Corpus} by \citet{Amsterdam} provides extensive manually annotated text that allows metaphor recognition in natural language for metaphor processing in English. It is frequently used to assess and train systems that need to recognise metaphorical usage on a large scale. Moreover, \citet{EPIE} presented a more condensed English idiom-oriented dataset with an emphasis on modelling idiomatic phrases as evaluative targets. It is typically employed to determine whether representations capture the conventionalised meanings underlying idioms or address them compositionally. The benchmark in \citet{stowe-etal-2022-impli} utilises paired instances to recast figurative understanding into a controlled evaluation format through the combination of a large semi-automatic section with a smaller manually selected gold set. Instead of focusing on the use of surface-level clues, it is meant to assess how effectively models handle figurative meaning, such as idioms and metaphors. 

The dataset by \citet{Reddy} provides human judgments on the transparency of a compound's meaning and the strength of its components' contributions. This dataset serves as a common baseline for forecasting noun compounds' levels of (non-)compositionality. \citet{liu-hwa-2016-phrasal} presents evaluation material for phrase-level robustness and rewriting where systems have to maintain meaning despite phrase replacements. This is helpful for evaluating phrase semantics and idiom-aware paraphrasing. %
In addition, \texttt{CoAM} by \citet{coam} focuses on the behaviour of multiword expressions in English and supports identification studies in which \textit{Multi-word Expressions} (MWEs) need to be regarded as single lexicalised components instead of distinct words. Furthermore, a number of English-only idiom benchmarks focus specifically on evaluating idiom competence rather than building linked lexical resources. Notable examples include \texttt{IDEM} by \citet{IDEM}, \texttt{IDIOMEM} by \citet{Haviv}, and \texttt{SLIDE} by \citet{SLIDE}.  With the objective to facilitate benchmarking and descriptive linguistic analysis in Danish, \texttt{The Danish Idiom Dataset} provides a selective collection of idioms and fixed expressions \cite{sorensen-nimb-2025-danish}. Swedish resources enhance this idiom-specific focus by extending coverage to MWEs more broadly. This allows for wider-coverage modelling of formulaic language and provides annotated material for recognising lexicalised MWEs beyond idioms \cite{Kurfal}. 
Furthermore, Germanic-language research frequently interacts with translation evaluation, especially in English-German contexts where specific idiom translation data allows for the methodical evaluation of MT/LLM errors such as literalization, semantic drift, and attenuation of figurative meaning during translation \cite{fadaee-etal-2018-examining}.

\subsection{Indic-Language Corpora}

\texttt{The Idiom Handling Dataset for Indian Languages} by \citet{agrawal-etal-2018-beating} provides idiom processing across several Indic languages such as Hindi, Urdu, Bengali,
Tamil, Gujarati, Malayalam,
Telugu, and typically includes mappings that enable cross-lingual handling, extending the coverage in Indic languages. In low-resource contexts, multilingual assessment and comparative analysis are enabled by \cite{agrawal-etal-2018-beating}.

In addition, the dataset presented by \cite{Singh} focuses on Hindi and Marathi idioms/MWEs within Indic languages, offering annotated content for MWE/idiom recognition and mode ling in these languages. \texttt{Konidiom} by\cite{Konidioms} provides idiom data for Konkani, a smaller, language-specific idiom resource that supports idiom research and resource development in a low-resource environment. 

\citet{Muhammad}'s dataset for Urdu focuses on translating idioms from Urdu and Roman Urdu, utilising idiom-focused test material to assess whether modern structures can preserve idiomatic meaning across script and language diversity. This is primarily an evaluation resource for translation behaviour under idiomaticity.

\subsection{Romance-Language Corpora}

Romance-language resources support a coherent discussion of how figurative meaning is represented within closely related languages and how well models transfer across them. By providing naturally grounded instances that allow idiom detection and interpretation in practical circumstances, \texttt{VIDiom-PT } supports this viewpoint for European Portuguese \cite{antunes-etal-2025-european}. In contrast, \texttt{Prometheus} emphasises meaning recovery at the discourse level and is proverb-oriented, making it simpler to comprehend multilingual proverbs through English–Italian data. By allowing systematic comparison between related Romance languages, standardised multilingual assessment strengthens these language-specific techniques. \texttt{SemEval-2022 Task 2} provides a common benchmark for English, Portuguese, and Galician in similar language circumstances, allowing for controlled assessment of cross-lingual generalisation and transfer \cite{madabushi2022semeval}.

\subsection{Cross-Lingual Figurative Language Corpora}
The large-scale cloze benchmark \texttt{ChID} by \citet{ChID} is employed to evaluate idiom comprehension in Chinese resources. It requires models to select a suitable idiom to fill in a passage's blank. In addition to testing contextual idiom understanding through blank-filling. In addition to assessing contextual idiom comprehending by blank-filling, the \texttt{Chengyu Cloze Test Dataset} by ~\citet{Jiang} emphasises semantic fit and discourse compatibility and delivers an invaluable, nearly equivalent evaluation environment.

Moreover, \texttt{PETCI} by ~\cite{PETCI} provides Chinese idioms related to English translations, facilitating the assessment of whether MT/LLM systems retain idiomatic meaning instead of generating literal, word-by-word renditions. Given this, it is extremely beneficial for controlled idiom translation testing.
By enabling idiom identification as well as analysis in morphosyntactically rich contexts, where inflexion and flexible surface forms can complicate detection and interpretation, Slavic-language corpora expand figurative language study beyond English ~\cite{aharodnik-etal-2018-designing, ParaDiom}. In order to allow both proverb retrieval/analysis and computational metaphor identification in a non-English setting, Greek corpora usually integrate structured proverb repositories with metaphor-annotated datasets \cite{Pavlopoulos,garcia-etal-2021-probing}. Through Hebrew and Arabic resources which facilitate MWE identification and metaphor detection in domain-specific contexts, including historically and stylistically unique texts that present additional model transfer challenges, Semitic corpora broaden coverage \cite{Liebeskind,Toker,Israa}. 

As a way to improve cross-lingual mapping and interoperability, multilingual linked idiom resources represent idioms as interconnected lexical entities across languages and link them to external lexical-semantic inventories \cite{LIDIOMS}. Furthermore, multilingual shared benchmarks support systematic analysis of cross-lingual generalisation and provide consistent comparison of systems on MWEs, idiomaticity, and phrase-level semantics through providing standardised annotation guidelines and evaluation protocols across various languages ~\cite{savary-etal-2023-parseme,semeval2013,madabushi2022semeval,tedeschi}. A summary of existing corpora, indicating the languages covered and the  FoS addressed in the above studies, is shown in Table~\ref{tab:datasets}.


\begin{table*}[!htb] 
\centering

\resizebox{!}{0.38\textheight}{
\begin{tabular}{|L{8cm}|C{5.0cm}|C{5cm}|} 

\hline
\multirow{2}{*}{\textbf{Dataset}}  & \multirow{2}{*}{\textbf{Languages}} &
\multirow{2}{*}{\textbf{FOS Explored}} \\ 
& & \\ \hline
IDIX~\cite{sporlederidioms} & English & Idioms \\
\hline
MAGPIE\cite{haagsma-etal-2020-magpie} & English &  Potentially Idiomatic Expressions \\
\hline
PIE~\cite{zhou-etal-2021-pie} & English &  Idiomatic Expressions (IE) \\ \hline
PIE(BNC and UKWaC)~\cite{adewumi-etal-2022-potential} & English & Metaphor, simile, euphemism, parallelism, personification, oxymoron, paradox, hyperbole, irony, and literal \\\hline
MABL~\cite{kabra-etal-2023-multi} & Hindi, Indonesian, Javanese, Kannada, Sundanese, Swahili and Yoruba &  Figurative language  \\\hline
VIDiom-PT~\cite{antunes-etal-2025-european} & European Portuguese & Verbal Idioms \\\hline
The Danish Idiom Dataset~\cite{sorensen-nimb-2025-danish} & Danish & Idiomatic expressions and fixed expressions \\\hline
LIDIOMS, DBnary,BabelNet~\cite{LIDIOMS} & English, German, Italian, Portuguese, and Russian & Idioms \\\hline
Prometheus~\cite{ozbal-etal-2016-prometheus} & English, Italian &Proverbs \\\hline
VU Amsterdam Metaphor Corpus~\cite{Amsterdam} & English & Metaphors \\\hline
MetaNet~\cite{dodge-etal-2015-metanet} & English, Russian, Mexican Spanish, Iranian Farsi &  Metaphors \\\hline
EPIE~\cite{EPIE} & English &Idiomatic Expressions \\\hline
IMPLI~\cite{stowe-etal-2022-impli} & English & Idiom, Metaphor \\\hline
ePiC~\cite{ePiC} & English & Proverbs \\\hline
ChID~\cite{ChID} & Chinese & Metaphor, Near-synonymy \\\hline
UPD*\cite{Reddy} & English & Compound Nouns \\\hline
SemEval-2013 Task 5 Dataset~\cite{semeval2013} & English, German, Italian & Phrases \\\hline
IdiomKB~\cite{li2024idiomkb} & English, Chinese, Japanese & Idioms \\\hline
IDEM ~\cite{IDEM} & English & Idioms \\\hline
IDIOMEM.~\cite{Haviv} & English & Idioms \\\hline
ID10M~\cite{tedeschi} & English, Chinese, Spanish, Dutch, French, German, Italian, Japanese, Polish, Portuguese & Idioms \\\hline
PETCI~\cite{PETCI} & Chinese, English &  Idioms \\\hline
AStitchInLanguageModels Dataset~\cite{AStitchInLanguageM} & English, Portuguese & Idioms \\\hline
UPD*~\cite{garcia-etal-2021-probing} & English & Idioms \\\hline
UPD*~\cite{Cordeiro} & English & Nominal Compounds \\\hline
SLIDE~\cite{SLIDE} & English & Idioms \\\hline
Russian Idiom-Annotated Corpus~\cite{aharodnik-etal-2018-designing} & Russian & Idiom \\\hline
UPD*\cite{fadaee-etal-2018-examining} & English, German & Idioms,Idiom Translation Dataset \\\hline
Idiom Handling Dataset for Indian Languages~\cite{agrawal-etal-2018-beating} & English, Hindi, Urdu, Bengali, Tamil, Gujarati, Malayalam, Telugu & Idioms \\\hline
Chengyu Cloze Test Dataset~\cite{Jiang} & Chinese & Idioms \\\hline
Multilingual Lexicon of Nominal Compound Compositionality~\cite{Ramisch} & English, French, Portuguese & Nominal Compounds \\\hline
UPD*~\cite{Pershina} & English,Idioms & Idiom Paraphrase Dataset \\\hline
Phrasal Substitution Dataset~\cite{Liu} & English & Idiomatic Expressions \\\hline
CoAM~\cite{coam} & English & MWEs \\\hline
ParaDiom~\cite{ParaDiom} & Slovene, English & Idiomatic Texts \\\hline
Konidioms Corpus~\cite{Konidioms} & Konkani & Idioms \\\hline
Multi-word Expression Dataset for Swedish~\cite{Kurfal} & Swedish & Multi-word Expression \\\hline
PARSEME Corpus Release 1.3 (VMWEs) \cite{savary-etal-2023-parseme} & Arabic, Bulgarian, Chinese, Croatian, Greek, Hebrew, Hindi, Irish, Latvian, Lithuanian, Maltese, Slovene, Turkish & Idioms, multiword expressions (verbal MWEs)\\\hline
SemEval-2022 Task 2 Dataset \cite{madabushi2022semeval}  & English, Portuguese, Galician & Idioms\\ \hline
UPD*\cite{Singh} & Hindi, Marathi & Idioms, MWEs\\ \hline
UPD* \cite{Liebeskind} & Hebrew & MWEs (incl. idiom-like fixed expressions)\\ \hline
Greek Proverb Atlas\cite{Pavlopoulos} & Greek & Proverbs\\ \hline
UPD* \cite{Florou} & Greek & Metaphor\\ \hline
UPD* \cite{Toker} & Hebrew & Metaphor\\ \hline
UPD* \cite{Muhammad} & Urdu, Roman Urdu & Idioms\\ \hline
AMC \cite{Israa} & Arabic & Metaphor\\ \hline

\end{tabular}}

\caption{ Existing Datasets Summary. *Corpora named `UPD' represent the \textit{Unnamed Primary Dataset(s)}, which includes papers that have released/utilised datasets without specific names.}
\label{tab:datasets}
\end{table*}

\section{Classification of Sinhalese Proverbs} \label{app:ClassificationProverb}

Here we discuss the classification of Sinhalese proverbs based on different criteria as given below.

\vspace{0.5em}
\noindent
\subsection{By the Nature of the Message (The Shape of the Message)}

\vspace{-0.2em}
\noindent

\paragraph{\hspace*{-4pt}
\raisebox{-0.5ex}{
\includegraphics[height=1.5\fontcharht\font`A]{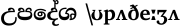}
}:} Proverbs that contain a moral lesson or advice. While not all proverbs are adages, some are interchangeably used to provide direct guidance, such as ``Don't burn your hand while the tongs are there''.

\vspace{-0.2em}
\noindent
\paragraph {\hspace*{-4pt}
\raisebox{-0.5ex}{ 
\includegraphics[height=1.5\fontcharht\font`A]{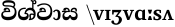}}:} Proverbs that express a commonly accepted social truth or collective belief rather than a direct instruction. These are sometimes referred to as ``Truth-principle proverbs'' (Sathyadharma Pirulu). Examples include ``A barking dog does not bite'' or ``Like eating the ear while sitting on the horn''.

\vspace{0.5em}
\noindent
\subsection{By Background Source (The Origin)}
      
\vspace{-0.2em}
\noindent\paragraph{\hspace*{-8pt}
\hspace*{-3pt}\raisebox{-0.5ex}{ 
\includegraphics[height=1.3 \fontcharht\font`A]{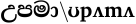} 
}:}  Ends in comparative markers. (\noindent\hspace*{-3pt}\raisebox{-0.5ex}{
\includegraphics[height=1.4\fontcharht\font`\A]{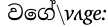}}, \raisebox{-0.5ex}{\includegraphics[height=1.4\fontcharht\font`\A]{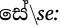}},\raisebox{-0.5ex}{
\includegraphics[height=1.4\fontcharht\font`\A]{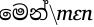} 
},\raisebox{-0.5ex}{
\includegraphics[height=1.4\fontcharht\font`\A]{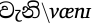} 
}).

\vspace{0.4em}
\noindent{\hspace*{-3pt}\raisebox{-0.5ex}{ \includegraphics[height=1.5\fontcharht\font`\A]{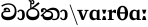}}:} \hspace*{-1pt}Ends in hearsay markers (\raisebox{-0.6ex}{ 
\includegraphics[height=1.2\fontcharht\font`\A]{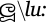}}).

\vspace{ 0.4em}
\noindent
\textbf{\hspace*{-3pt}\raisebox{-0.4ex}{ 
\includegraphics[height=1.4\fontcharht\font`\A]{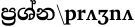} 
}:} \hspace*{-15pt}Ends in interrogative markers (\raisebox{-0.5ex}{ 
\includegraphics[height=1.3\fontcharht\font`\A]{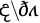} 
}), often acting as rhetorical devices to prompt self-reflection (e.g.,\hspace*{-1pt}``\raisebox{-0.5ex}{ 
\includegraphics[height=1.56\fontcharht\font`\A]{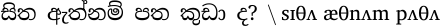} 
}
\hspace*{-3pt}\raisebox{-0.5ex}{ 
\includegraphics[height=1\fontcharht\font`\A]{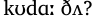}''
}
    
\vspace{0.5em}
\noindent
\raisebox{-0.4ex}{ 
\includegraphics[height=1.5\fontcharht\font`\A]{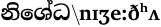} 
}(Negative): Ends in negation. (\raisebox{-0.5ex}{ 
\includegraphics[height=1.3\fontcharht\font`\A]{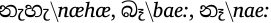}}).

\vspace{0.5em}
\noindent
\subsection{By Grammatical Ending (The Marker)}

\vspace{0.5em}
\noindent
\raisebox{-0.5ex}{%
\includegraphics[height=1.4\fontcharht\font`\A,page=1]{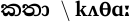}
} \textbf{(Story-based):} These proverbs rely on shared cultural memory. They are often unintelligible without knowledge of the specific folktale or historical event (e.g., ``\raisebox{-0.5ex}{%
  \includegraphics[height=1.5\fontcharht\font`\A,page=1]{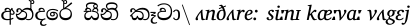} 
}'' - Like Andare eating sugar).

\vspace{0.5em}
\noindent
\raisebox{-0.5ex}{%
\includegraphics[height=1.4\fontcharht\font`\A,page=1]{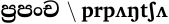}
}\textbf{ (Phenomenon-based):} These are derived from empirical observations of the agrarian environment, nature, or daily life (e.g., ``\raisebox{-0.5ex}{%
\includegraphics[height=1.6\fontcharht\font`\A,page=1]{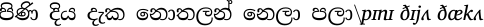}
} \raisebox{-0.5ex}{%
\includegraphics[height=1.3\fontcharht\font`\A,page=1]{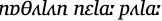}
}'' - Do not crush the greens, seeing the dew).

\vspace{0.5em}
\noindent
\raisebox{-0.5ex}{%
  \includegraphics[height=1.5\fontcharht\font`\A,page=1]{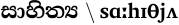}
}\textbf{(Literature-based):} These originate from classical texts such as the \raisebox{-0.5ex}{%
  \includegraphics[height=1.5\fontcharht\font`\A,page=1]{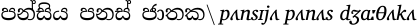}
} \hspace*{-3pt}or \raisebox{-0.5ex}{%
  \includegraphics[height=1.5\fontcharht\font`\A,page=1]{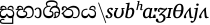}
}\hspace*{-3pt}, reflecting the influence of Buddhism and literacy on folk speech.

\vspace{0.5em}
\noindent
Among these, \raisebox{-0.5ex}{ 
    \includegraphics[height=1.3\fontcharht\font`\A]{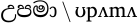} 
    } (Simile) sub-category is the most prevalent. This indicates that analogical reasoning, understanding one concept in terms of another, is the primary cognitive tool used in Sinhala folk wisdom.
\raisebox{-0.5ex}{
    \includegraphics[height=1.5\fontcharht\font`\A]{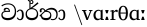} 
    } (Report) category is the second most common proverb structure. The prevalence of the particle \raisebox{-0.5ex}{ 
    \includegraphics[height=1.5\fontcharht\font`\A]{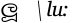} 
    } (it is said) underscores the importance of oral tradition and collective knowledge in Sri Lankan culture, wisdom is validated not by the speaker's authority, but by the fact that ``it has been said'' by ancestors.

\section{Sinhala Proverbs vs Sinhala Idioms}
\label{appendix:strategies}

\paragraph{The Dichotomy of Sinhala Proverbs and Sinhala Idioms:} While both categories function as figurative devices, they are distinguishable through three primary dimensions: Syntactic Structure, Semantic Deductibility, and Pragmatic Function.
\paragraph{Semantic Deductibility (Opacity vs. Transparency):} Idioms in Sinhala often exhibit high semantic opacity; a learner cannot easily deduce that ``\textit{\raisebox{-0.5ex}{
    \includegraphics[height=1.3\fontcharht\font`\A]{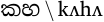}}}'' in ``\textit{\raisebox{-0.5ex}{
    \includegraphics[height=1.5\fontcharht\font`\A]{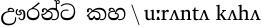}}}'' implies ``wasting resources.'' However, Proverbs are often semantically translucent. Even a first-time listener can deduce the meaning of ``\textit{\raisebox{-0.5ex}{
    \includegraphics[height=1.5\fontcharht\font`\A]{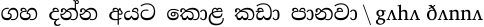}} \hspace*{-3pt}\raisebox{-0.5ex}{
    \includegraphics[height=1.3\fontcharht\font`\A]{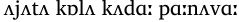}}}'' (Showing leaves to those who know the tree) based on the imagery of deception and expertise.
\paragraph{Pragmatic Function:} \textit{\raisebox{-0.5ex}{
    \includegraphics[height=1.5\fontcharht\font`\A]{Figures/prasthapiruluwb.pdf}}} are didactic; they convey general truths, social beliefs, or moral advice (\textit{\raisebox{-0.5ex}{
    \includegraphics[height=1.5\fontcharht\font`\A]{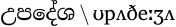}}}). 
    \textit{\hspace*{-3pt}\raisebox{-0.5ex}{
    \includegraphics[height=1.5\fontcharht\font`\A]{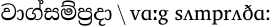}}} are descriptive; they categorise a state of being or an action without necessarily offering a moral judgment.
\paragraph{Dominance of Idioms:}{\raisebox{-0.4ex}{
\includegraphics[height=1.4\fontcharht\font`\A]{Figures/wagsamprada.pdf}}} constitute the overwhelming majority of the dataset. This quantitative dominance suggests that Sinhala speakers prioritise ``descriptive efficiency'' in daily language, using short, culturally loaded phrases to quickly describe complex situations, over the more formal, structured wisdom of proverbs.

\section{Dataset Annotation}
\label{app:dataset}

\begin{figure}
    \centering
    \setlength{\fboxsep}{1pt} 
    \setlength{\fboxrule}{0.4pt} 
    \fbox{%
        \includegraphics[width=0.9\columnwidth]{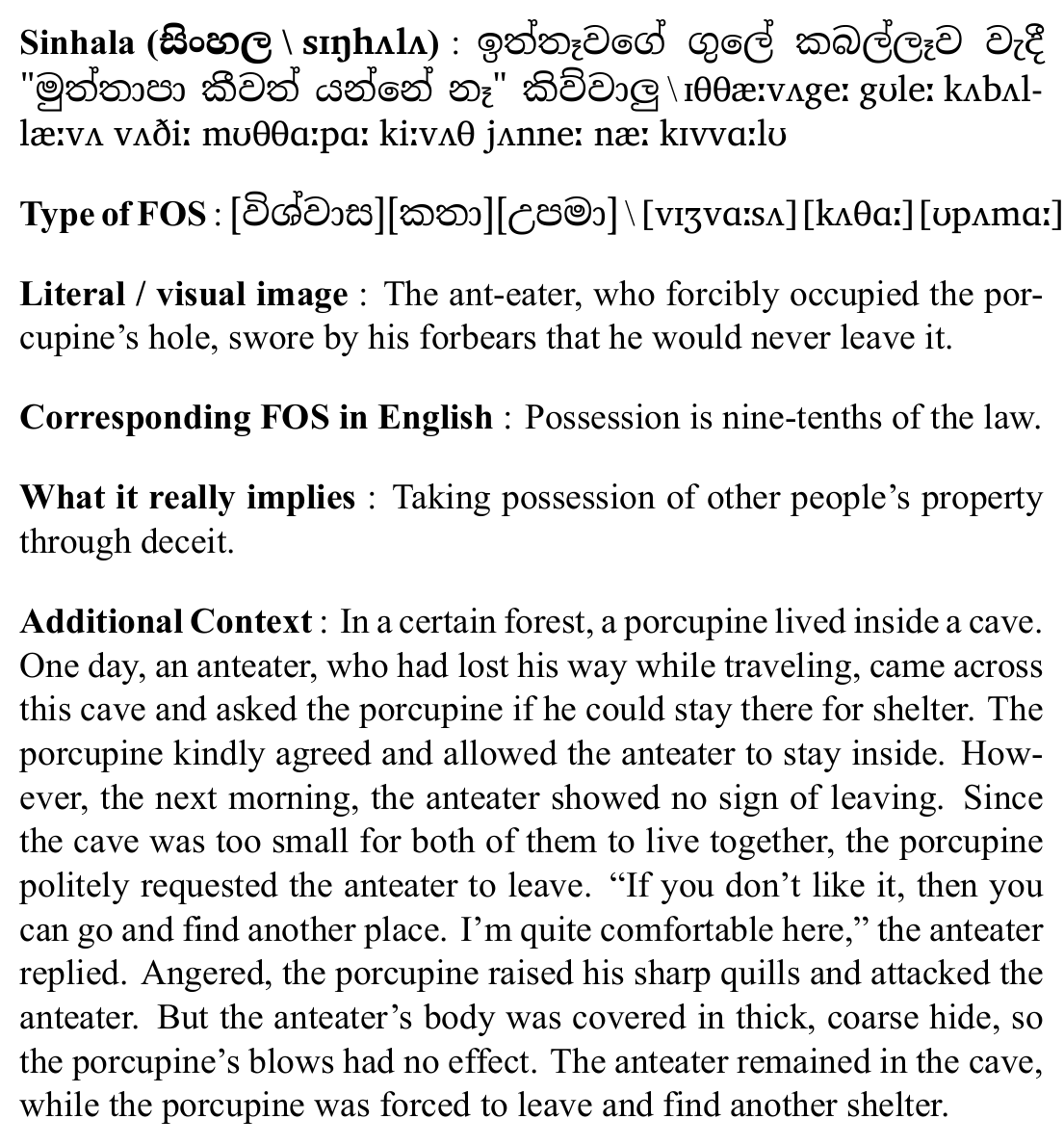}%
    }%
    \caption{An example of a record on the dataset.}
    \label{fig:record}
\end{figure}

The dataset was annotated by filling in the fields. Not all fields were filled in for all records, as shown in Table~\ref{tab:field_counts}. Figure \ref{fig:record} contains an example of a record in the dataset.

\section{Cultural Analysis}
\label{app:cultural_analysis}

The Source Domain explores the abstract imagery and objects used in the FoS to deliver the message, whilst the target theme is used to identify the messages delivered by the various FoS.


\subsection{Specific Cultural Codes}

Certain symbols carry specific, unchangeable meanings in the Sinhala cultural lexicon. The following are some of the examples utilised in Sinhala FoS.

\vspace{0.5em}
\noindent
\textbf{The Elephant (Power \& Scale):} The elephant is the cultural yardstick for greatness. It is used to contrast ``the great'' with ``the small.'' It represents forces that are often too big to manage or criticise.

\vspace{0.5em}
\noindent
\textbf{The Dog (Low Status):} In contrast to the elephant, the dog is consistently used to represent unworthiness or low social status. It serves as a warning of what happens when one lacks dignity.

\vspace{0.5em}
\noindent
\textbf{The Tree (Character):} Trees are almost always metaphors for moral character. A person is judged like a tree, by their ``fruit'' (utility to society) or their ``wood'' (strength/weakness).

\subsection{Emotional Landscape}

The sentiment analysis shows that the vast majority of FoS (83\% of the data) are \textit{Neutral}. They are not optimistic or pessimistic; they are descriptive. The culture does not say ``Life is good'' or ``Life is bad''; it says, ``If you take this action, the corresponding outcome will occur inevitably.'' It values truth over comfort.

Table \ref{tab:domain_occurrences} provides a comprehensive overview of the cultural analysis, summarising the frequency of literal imagery and the specific thematic domains explored within the \textsc{SinFoS} dataset. 

\begin{table}[h!tb]
\centering
\resizebox{0.4\textwidth}{!}{
\begin{tabular}{l r}
\hline
\textbf{Category} & \textbf{No. of Occurrences} \\
\hline
\multicolumn{2}{l}{\textit{Source Domain (Literal Imagery)}} \\
Body \& Senses (Somatic) & 242 \\
Nature \& Agriculture & 194 \\
Animals (Fauna) & 148 \\
Household \& Daily Life & 144 \\
\hline
\multicolumn{2}{l}{\textit{Target Theme (Cultural Meaning)}} \\
Ethics \& Moral Character & 162 \\
Karma \& Consequence & 127 \\
Impermanence \& Uncertainty & 121 \\
Social Status \& Hierarchy & 73 \\
Human Relations \& Conflict & 64 \\
\hline
\end{tabular}}
\caption{Distribution of literal source domains and abstract cultural themes observed in the \textsc{SinFoS} dataset via hybrid thematic analysis.}
\label{tab:domain_occurrences}
\end{table}

\section{Model Classification}
\label{class_performance}

The results of classifying proverbs and idioms are summarised in Table~\ref{tab:model_performance_1}. Word2Vec showed the best performance for Naive-Bayes and Linear SVC in terms of recall and accuracy. In contrast, TF-IDF 3-gram vectorisation excelled with Random Forest, XGBoost, and the ensemble model combining these with Linear SVC.

\begin{table}[h!tb]
\centering
\resizebox{0.32\textwidth}{!}{
\begin{tabular}{|p{3.5cm}|c|c|c|}
\hline
\textbf{Experiments} & \textbf{Acc.} & \textbf{P-Rec.} & \textbf{I-Rec.} \\ \hline
TF-IDF (Unigram) Multimodal Naive-Bayes & 0.7167 & 0.59 & 0.81  \\ \hline
TF-IDF (Char 3-gram) Multimodal Naive-Bayes & 0.8348 & 0.79 & 0.87  \\ \hline
Gaussian Naive-Bayes Word2Vec & 0.9034 & 0.92 & 0.89  \\ \hline
TF-IDF (Unigram) Linear SVC & 0.7639 & 0.64 & 0.86  \\ \hline
TF-IDF (Char 3-gram) Linear SVC & 0.8712 & 0.81 & 0.92  \\ \hline
Linear SVC Word2Vec & 0.9034 & 0.90 & 0.91  \\ \hline
Random Forest (tuning) TF-IDF (Unigram) & 0.8026 & 0.66 & 0.91  \\ \hline
Random Forest (tuning) TF-IDF (Char 3-gram) & 0.8927 & 0.83 & 0.94  \\ \hline
Random Forest (tuning) Word2Vec & 0.8755 & 0.81 & 0.93  \\ \hline
XGBoost (tuning) TF-IDF (Unigram) & 0.8090 & 0.65 & 0.93  \\ \hline
XGBoost (tuning) TF-IDF (Char 3-gram) & 0.9013 & 0.86 & 0.93  \\ \hline
XGBoost (tuning) Word2Vec & 0.8690 & 0.83 & 0.90  \\ \hline
TF-IDF (Char 3-gram) Voting Ensemble & 0.9056 & 0.85 & 0.95  \\ \hline
Voting Ensemble Word2Vec & 0.8884 & 0.85 & 0.92  \\ \hline
Bi-LSTM & 0.9163 & 0.86 & 0.96  \\ \hline
Deep NN & 0.9270 & 0.94 & 0.92  \\ \hline
\end{tabular}}
\caption{Model Performance: Accuracy, Proverbs Recall (P-Rec.), and Idioms Recall (I-Rec.).}
\label{tab:model_performance_1}
\end{table}

\section{Performance of all LLMs}
\label{llm_performance}

A brief overview of each metric's blind spots and how each metric mitigates the other's is provided in Table \ref{tab:metrics}.

\begin{table}[h!tb]
    \centering
    \footnotesize 
    \renewcommand{\arraystretch}{1.4}
    \resizebox{0.35\textwidth}{!}{
    \begin{tabular}{@{} p{0.18\columnwidth} | p{0.38\columnwidth}| p{0.38\columnwidth} @{}}
        \hline
        \textbf{Metric} & \textbf{Blind Spot} & \textbf{Mitigation Strategy} \\
        \hline
        Cosine Similarity & 
        The ``Keyword Bag'' Problem: the model may achieve high scores by guessing relevant keywords even if the grammatical structure is flawed. & 
        Fidelity acts as a ``Logic Gate,'' requiring semantic validity rather than just keyword overlap. \\
        \hline
        Fidelity Score & 
        The ``Hyper-Literal'' Problem: creative paraphrases with different structures might be penalised. & 
        Cosine Similarity permits creative phrasing; high similarity with low fidelity suggests a valid non-standard translation. \\
        \hline
    \end{tabular}}
    \caption{Evaluation Metrics and Mitigation of their Blind Spots}
    \label{tab:metrics}
\end{table}

\begin{figure}[ht]
    \centering
    \includegraphics[width=0.9\linewidth]{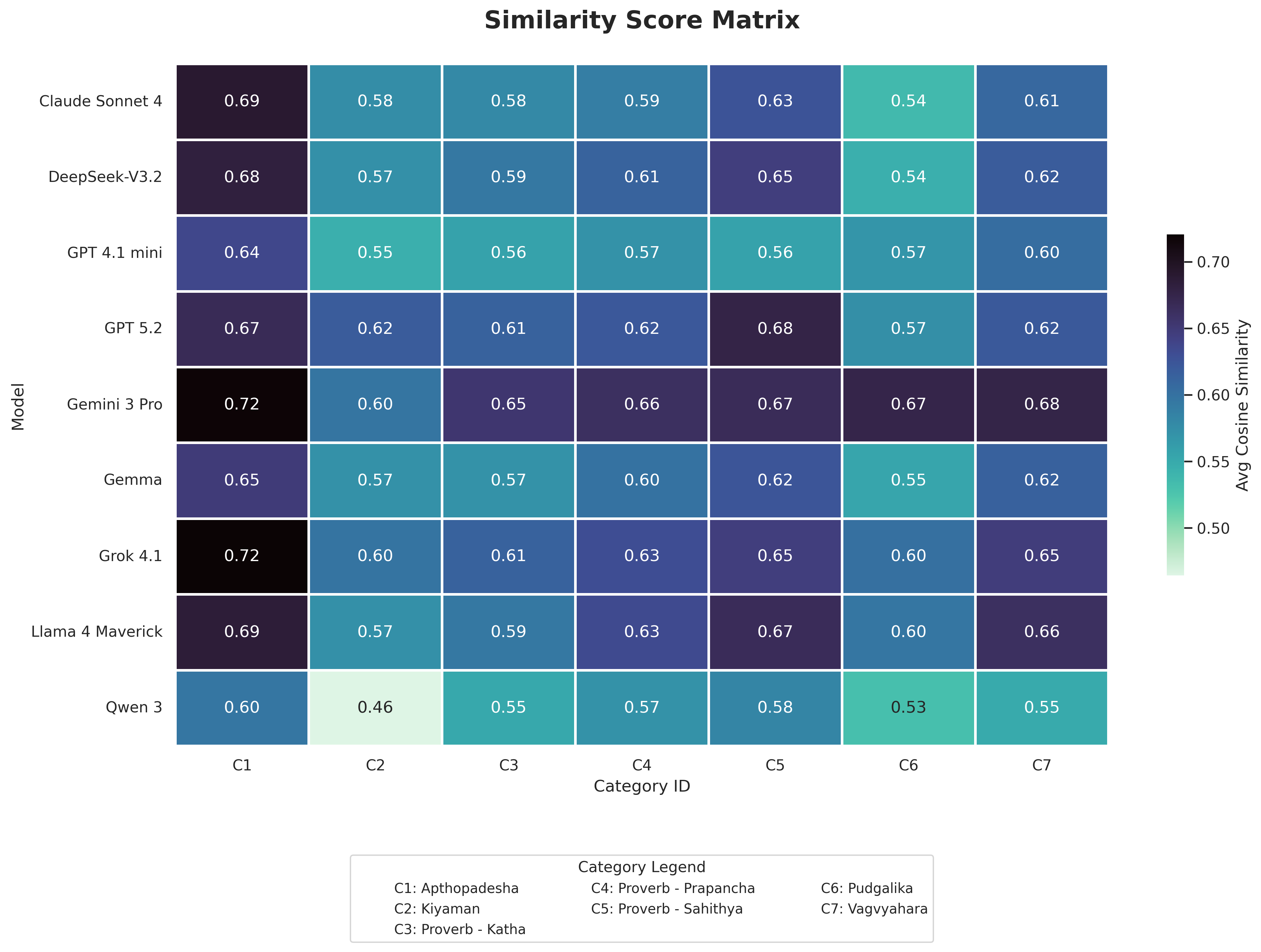}
    \caption{Benchmarking LLM Performance: Cosine Similarity.}
    \label{fig:similarity_1}
\end{figure}

\begin{figure}[ht]
    \centering
    \includegraphics[width=0.9\linewidth]{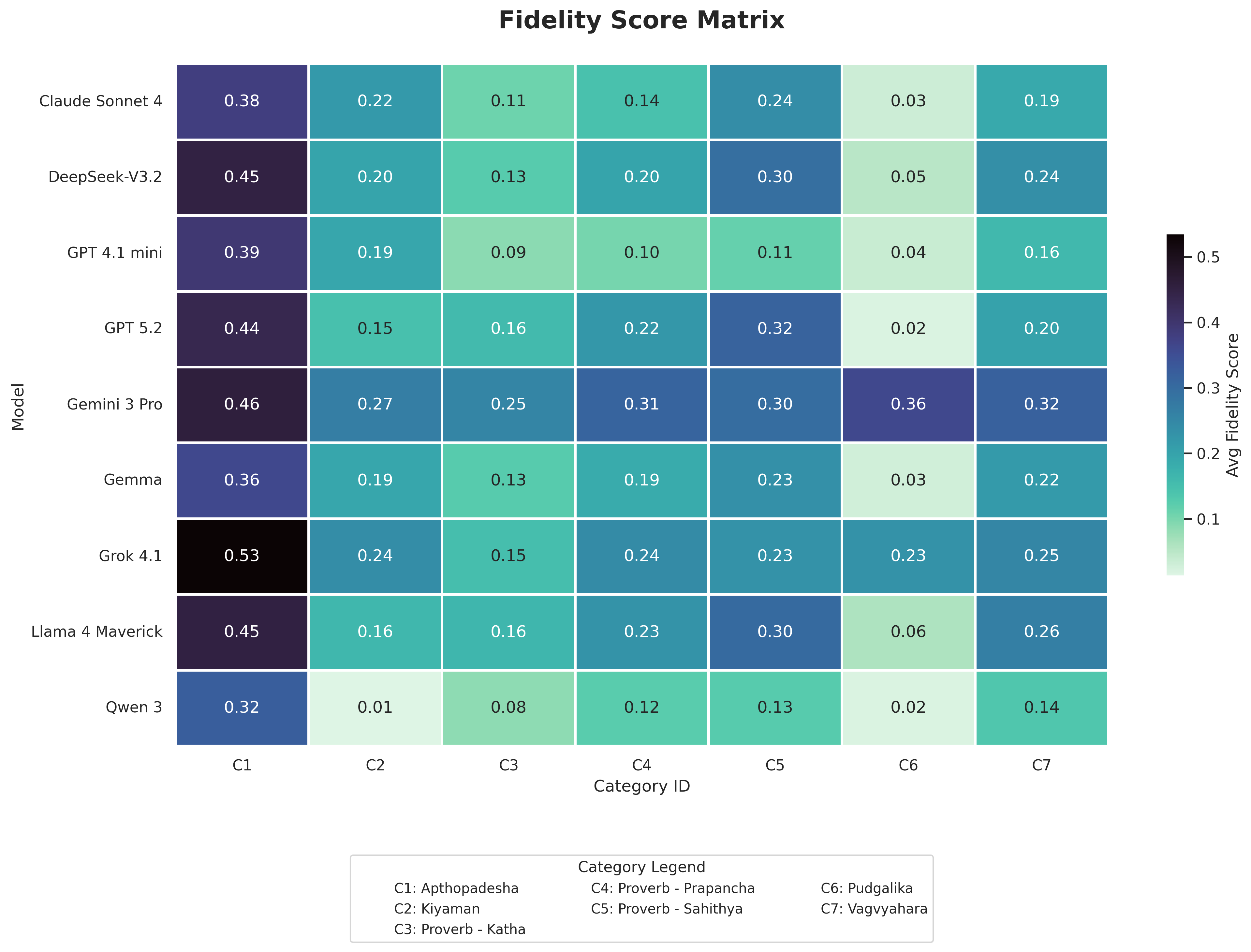}
    \caption{Benchmarking LLM Performance: Fidelity Scores.}
    \label{fig:fidelity_1}
\end{figure}

The Figures \ref{fig:similarity_1} and \ref{fig:fidelity_1} represent the Cosine Similarity scores and Fidelity Scores of all the models across seven different categories. Along with \raisebox{-0.5ex}{\includegraphics[height=1.5\fontcharht\font`\A]{Figures/apthopadeshawb.pdf}}, the models seem to have decent performances for proverbs associated with nature as they seem to be able to decipher the meaning using the phenomenon. In the case of \raisebox{-0.5ex}{\includegraphics[height=1.5\fontcharht\font`\A]{Figures/kiyamanwb.pdf}} though, as in proverbs based on folklore, the language models seem to struggle. This is tied with the fact that unlike \raisebox{-0.5ex}{\includegraphics[height=1.5\fontcharht\font`\A]{Figures/apthopadeshawb.pdf}}, \raisebox{-0.5ex}{\includegraphics[height=1.5\fontcharht\font`\A]{Figures/kiyamanwb.pdf}} are more specific to the language.

\end{document}